\documentclass{article} 
\usepackage{iclr2027_conference,times}

\usepackage{amsmath,amssymb}

\usepackage{amsmath,amsfonts,bm}

\def\eqref#1{equation~\ref{#1}}

\def\1{\bm{1}}

\DeclareMathAlphabet{\mathsfit}{\encodingdefault}{\sfdefault}{m}{sl}
\SetMathAlphabet{\mathsfit}{bold}{\encodingdefault}{\sfdefault}{bx}{n}

\usepackage{booktabs}
\usepackage{array}
\usepackage{multirow}
\usepackage{algorithm}
\usepackage{algpseudocode}
\usepackage{pifont}
\usepackage{xspace}
\usepackage{microtype}
\usepackage{enumitem}
\usepackage{xcolor}
\usepackage{hyperref}
\usepackage{url}
\usepackage{graphicx}

\usepackage{tikz}
\newcommand*\circled[1]{\tikz[baseline=(char.base)]{
        \node[shape=circle, draw, fill=white, 
              minimum size=1em, inner sep=0.2pt,
              text=black] (char) {#1};}}    
\newcommand*\circledB[1]{\tikz[baseline=(char.base)]{
            \node[shape=circle,fill,inner sep=0.2pt] (char) {\textcolor{white}{#1}};}}
\definecolor{myblue}{RGB}{15,82,186}

\usetikzlibrary{tikzmark}
\tikzset{circledColor/.style={circle,draw,inner sep=0.1em,line width=0.04em}}
 
\newcommand{\add}[1]{\textcolor{blue}{#1}}
\newcommand{\addx}[1]{\textcolor{red}{#1}}
\usepackage{enumitem}
\usepackage{multirow}
\usepackage{wrapfig}

\newcommand{\curezero}{RobustNCLv1\xspace}
\newcommand{\curemode}{RobustNCLv3\xspace}
\newcommand{\curemm}{RobustNCLv2\xspace}
\newcommand{\rate}[2]{$#1{\times}10^{-#2}$}

\setlist[itemize]{leftmargin=*,topsep=2pt,itemsep=1pt,parsep=0pt}

\title{RobustNCL: Robustifying Neuromorphic Continual Learning under Transient Faults with Lightweight Detect-and-Cure Strategy}

\author{Rachmad Vidya Wicaksana Putra, Shuochen Zhou, Muhammad Shafique\\
eBRAIN Lab, New York University (NYU) Abu Dhabi, United Arab Emirates\\
Email: \{rachmad.putra, sz4544, muhammad.shafique\}@nyu.edu}

\iclrfinalcopy 
\begin{document}

\maketitle

\begin{abstract}
In neural networks, continual learning (CL) is a prominent approach for enabling deployed systems to adapt to dynamic operational environments by effectively learning from new unseen data at runtime. 
Recently, the neuromorphic continual learning (NCL) concept has emerged as an energy-efficient alternative to CL by leveraging sparse event-based operations. 
However, NCL performance can be undermined by transient faults in the hardware platforms, which may come from high-energy particle strikes at any time. 
Currently, NCL reliability under transient faults has not been studied, and hence the corresponding fault-tolerance techniques remain unexplored.
To address this, we propose a novel RobustNCL framework to improve NCL reliability in the presence of transient faults in the DRAM-based off-chip memory of hardware platforms through a lightweight detect-and-cure strategy. 
Specifically, it first monitors the presence of faults during weight data access from off-chip memory, and then cures the accessed weights if anomalous signatures are detected. 
It provides three curing variants: replacement with zero, min/max value, and most-probable value from the original/clean weight distribution. 
To support this, the weight curing values are stored in the radiation-hardened memory. 
Experimental results show that, in the presence of high fault rates (e.g., $5{\times}10^{-4}$ for memory replay), RobustNCL maintains high accuracy close to the accuracy of fault-free models (i.e., within -0.5\% accuracy for the best RobustNCL variant) and significantly improves accuracy from the baseline without fault mitigation across different NCL methods and datasets (i.e., Split MNIST, Split Fashion-MNIST, Split CIFAR-10, and Split CIFAR-100), while incurring negligible latency and energy overheads. These results highlight that our RobustNCL framework is a promising approach for enabling reliable and energy-efficient NCL systems.
\end{abstract}


\begin{figure*}[h]
    \centering
    \includegraphics[width=\linewidth]{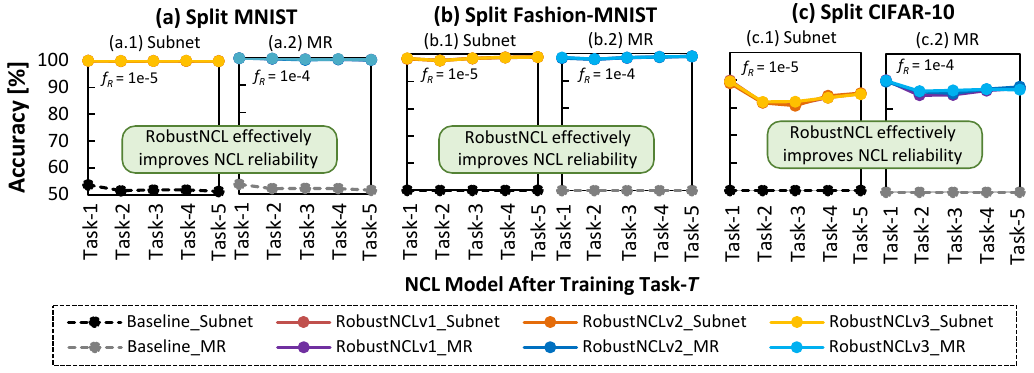}
    \caption{RobustNCL effectively improves NCL reliability againts transient faults in the DRAM off-chip/main memory under a task-incremental learning (TIL) scenario across different benchmarks: (a)~Split MNIST, (b)~Split Fashion-MNIST, and (c)~Split CIFAR-10, across different NCL methods: task-specific subnetwork (Subnet) and memory replay (MR). 
    The fault rate is denoted as $f_R$ considering a DDR3-1600 2Gb DRAM chip~\citep{Ref_Micron}, RobustNCL variants are with v1-v3 codes.}
    \label{Fig_Teaser}
    \vspace{-0.5cm}
\end{figure*}

\section{Introduction}
\label{Sec_Intro}

Neural network (NN)-based systems typically rely on offline training to gain high performance (e.g., accuracy) during deployments~\citep{Ref_LeCun_DeepLearning_Nature15}.
However, the systems' knowledge may be outdated over time since they encounter new/unseen data at runtime, that were not learned during offline training and often represent a new task. 
This condition negatively affects system performance, such as significant accuracy degradation.
Recently, emerging resource-constrained autonomous systems (e.g., mobile robots, UAVs, UGVs, etc.) increasingly demand that deployed systems can learn new tasks using new data gathered directly from dynamically changing environments under low computational and memory costs~\citep{Ref_Shaheen_CLsurvey_JIRS22}. 

To address this, the continual learning (CL) concept has become a prominent approach. 
CL allows deployed NN-based systems to learn new tasks from data encountered at runtime without retraining on all past training data from previously-learned tasks, while minimizing catastrophic forgetting (CF) issues, where the NN models forget older tasks when learning a new task~\citep{delange2022continual,wang2024comprehensive,minhas2025continual}. 
However, CL methods often demand high computational and memory costs, thereby posing non-trivial challenges for their deployments on resource-constrained autonomous systems.
Recently, the neuromorphic continual learning (NCL) concept has been explored to offer an energy-efficient alternative to conventional CL methods by leveraging event-driven, sparse spiking operations from spiking neural network (SNN) computational models~\citep{roy2019towards,davies2018loihi,minhas2025continual}. 

NCL accuracy can be undermined by \textit{transient faults} in hardware platforms, especially in the DRAM off-chip/main memory that stores the trained weights; see Fig~\ref{Fig_Teaser}. 
Transient faults can occur at any time due to high-energy particle strikes, from either cosmic rays or packaging materials~\citep{baumann2005radiation, putra2022softsnn}.
These faults manifest as bit-flips at the hardware layer, and these errors can propagate to the software layer~\citep{putra2022softsnn}.
These errors persist until they are replaced with new data bits, hence also known as \textit{soft errors}.
Such errors change the weight values in the DRAM at any time, and may lead to significant NCL accuracy degradation.
However, NCL reliability under transient faults has not been studied, and the corresponding fault-tolerance techniques also remain unexplored.
\textit{Therefore, \textbf{the targeted problem} in this paper is, how can we systematically study the negative impact of transient faults in the DRAM off-chip/main memory on NCL reliability, and develop an efficient fault-mitigation technique?}
An efficient solution to this problem may enable reliable NCL processing with low memory, latency, and energy overheads, which is especially suitable for resource-constrained autonomous systems.

\textbf{State-of-the-Art Fault Tolerance Methods:}
Standard fault-tolerance methods from integrated circuit (IC) design may be used to mitigate transient faults. 
They include a redundancy approach like Dual Modular Redundancy (DMR)~\citep{Ref_Vadlamani_DMR_DATE10} and Triple Modular Redundancy (TMR)~\citep{Ref_Lyons_TMR_IBM62}, and an error correction approach like Error Correcting Codes (ECCs)~\citep{Ref_Sze_ECCs_USPatent00}. 
However, these techniques incur high overheads in terms of area, latency, and power/energy consumption.
Additionally, fault-tolerance methods from offline-trained Deep Neural Networks (DNNs)~\cite{hoang2020ftclipact} and Spiking Neural Networks (SNNs)~\citep{putra2022softsnn} cannot be directly employed because they do not consider different states of the network model (i.e., updated models due to NCL process) that can be affected by transient faults.  
Furthermore, the state-of-the-art NCL works still focus on improving their performance for the targeted applications.
\textit{Therefore, NCL reliability against transient faults has not been studied, and consequently, the corresponding fault-mitigation techniques remain underexplored.}

\begin{figure*}[t]
    \centering
    \includegraphics[width=\linewidth]{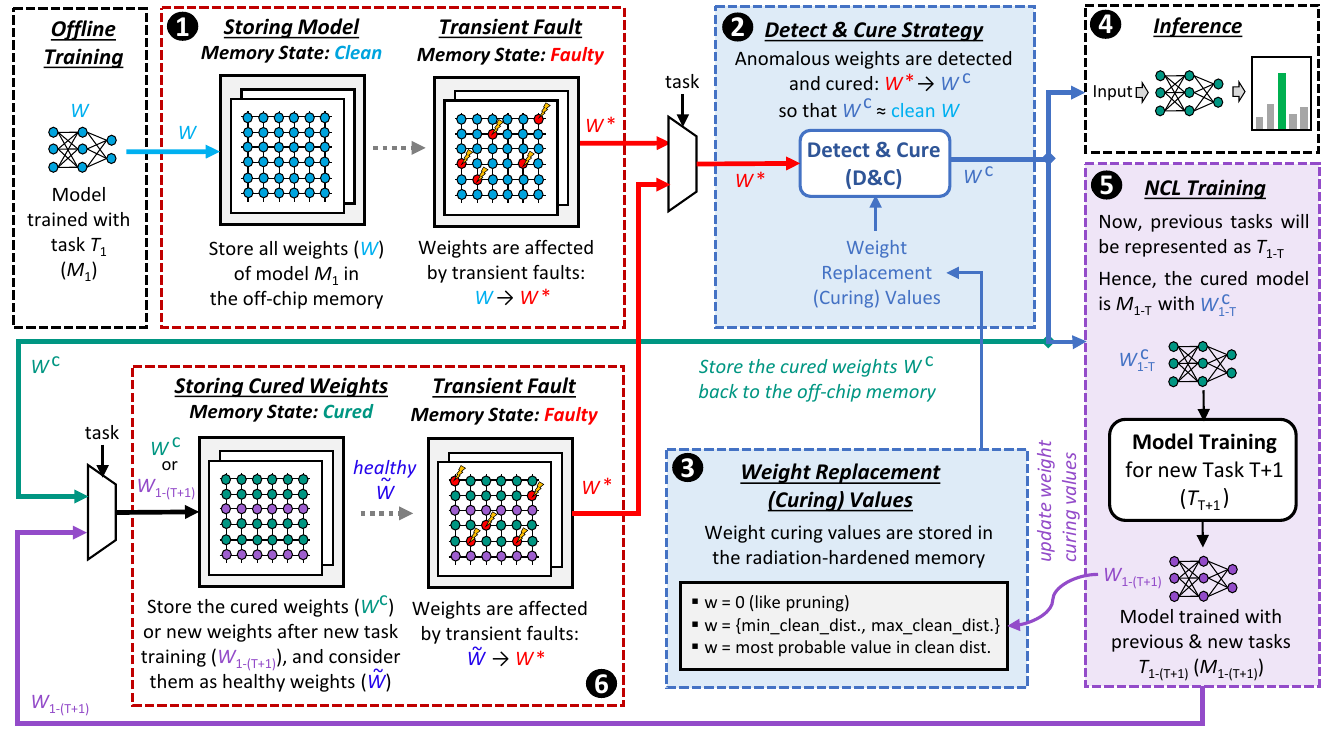}
    \caption{The RobustNCL framework, showing the transient fault exposure stage, our detect-and-cure (D\&C) strategy stage supported by weight replacement (curing) value selection, and the NCL training stage.
    RobustNCL variants depend on how the faulty weights are replaced by selected curing weights.}
    \label{Fig_RobustNCL}
\end{figure*}

\textbf{Our Novel Contributions:} 
To address the research problem, we propose \textit{\textbf{RobustNCL}, a novel framework that aims at \underline{Robust}ifying \underline{NCL} under transient faults in the DRAM off-chip/main memory with a lightweight detect-and-cure (D\&C) strategy}; see Fig.~\ref{Fig_RobustNCL}. 
This study is also the first work to investigate NCL reliability in the presence of hardware faults and propose an efficient fault-mitigation solution.
RobustNCL achieves this through the following novel contributions.
\begin{itemize} 
    \item \textbf{NCL Transient-Fault Modeling:} 
    It introduces a fault model for injecting transient faults in the NCL scenario, which is important and required for systematic study and analysis.
    Its key idea is applying transient faults to the DRAM at different possible points in time. 
    \item \textbf{NCL Fault-Tolerance Analysis:} 
    It systematically studies NCL fault tolerance and its characteristics (e.g., weight distributions) through experiments that investigate NCL performance/accuracy under transient faults in the DRAM across different fault rates ($f_R$) and different NCL methods. 
    \item \textbf{Lightweight Detect-and-Cure (D\&C) Strategy:} 
    It continuously detects anomalous weights that are accessed from the DRAM based on their values, and replaces (cures) them with weight values that lead to safe spiking activities based on observations in the NCL training phase. 
    It also provides appropriate curing weight variants, considering different possible model states due to NCL training. 
    \item \textbf{Comprehensive Evaluation:}
    We evaluate the RobustNCL framework in diverse settings based on combinations of different fault rates $f_R$, different benchmarking datasets, and different NCL methods: task-specific subnetwork (Subnet) and memory replay (MR). 
\end{itemize}

\section{Related Work}
\label{Sec_Prelim}

\begin{figure}[t]
    \centering
    \includegraphics[width=0.875\linewidth]{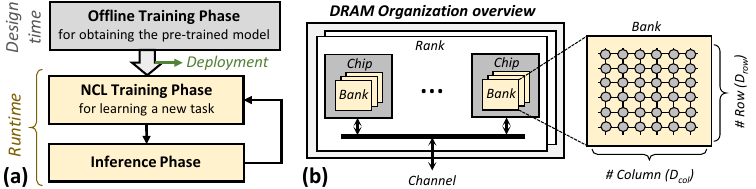}
    \caption{(a) The NCL pipeline consists of three phases: the offline training at design time, the NCL training at runtime, and the inference at runtime. (b) DRAM  organization overview, including channels, ranks, chips, and banks.}
    \label{Fig_NCL_DRAM}
\end{figure}

\textbf{Neuromorphic Continual Learning (NCL):} 
NCL combines the event-driven computation of SNNs~\citep{maass1997networks,roy2019towards} with continual learning~\citep{minhas2025continual}.
In general, an NCL pipeline consists of three phases: (1) the offline training at design time, (2) the NCL training at runtime, and (3) the inference at runtime; see Fig.~\ref{Fig_NCL_DRAM}(a).
At runtime, the deployed system will be in the NCL training phase when the network model learns a new task; otherwise, it is in the inference phase while employing the most updated model. 
NCL scenarios depend on the task identity.
In this paper, we consider the task incremental learning (TIL) scenario, as it is widely used as the standard NCL evaluation in the SNN community.
TIL has several characteristics: the task boundaries are explicitly defined, the task identity is known in both training and inference phases, and each task usually has multiple classes. 
In the TIL scenario, $\mathcal{T}_1,\dots,\mathcal{T}_K$ arrive sequentially, and each task has its own classification head $\mathbf{h}_t$. Let $a_{t,j}$ denote the test accuracy on $\mathcal{T}_j$ after learning $\mathcal{T}_t$. 
We consider the final average accuracy as $\mathrm{Acc}=\frac{1}{K}\sum_{j=1}^{K}a_{K,j}$~\citep{lopezpaz2017gradient}.

\begin{figure}[t]
    \centering
    \includegraphics[width=\linewidth]{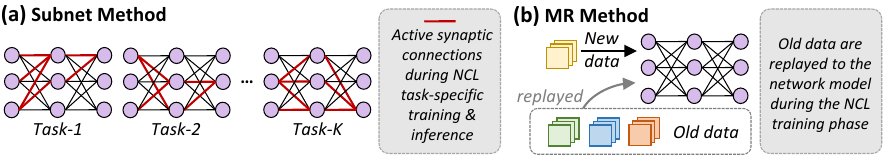}
    \caption{Overview of (a) the task-specific subnetwork \textit{(Subnet)} method and (b) the memory replay \textit{(MR)} method that are considered in this paper.}
    \label{Fig_ENCL_MR}
\end{figure}

NCL methods can be classified into several approaches, such as architecture methods~\citep{han2023dynamic, pes2024active}, memory replay methods~\citep{dequino2024compressed, minhas2025replay4ncl}, and learning enhancement methods~\citep{Ref_Putra_SpikeDyn_DAC21, Ref_Putra_lpSpikeCon_IJCNN22, skatchkovsky2022bayesian, antonov2022continuous, xiao2024hebbian}.
In this paper, we consider two NCL methods from different approaches: \textit{the task-specific subnetwork (Subnet) method} from the architecture-based approach, and \textit{the memory replay (MR) method} to show the generality of our RobustNCL framework. 
In the Subnet methods, a specific subnetwork is trained for a specific task, allocating a pathway within the network to process that task; see Fig.~\ref{Fig_ENCL_MR}(a).
They typically achieve competitive accuracy with relatively small memory and computational overheads, but they may face limitations related to representation capacity. 
Meanwhile, MR methods replay previous inputs to the network model to reinforce old knowledge; see Fig.~\ref{Fig_ENCL_MR}(b). 
Consequently, they often offer high accuracy but at the cost of a large replay buffer and extra forward/backward passes, indicating large memory and computational overheads~\citep{proietti2023memory, dequino2024compressed, minhas2025replay4ncl}. 

\textbf{DRAM Fundamentals:} 
The top-down hierarchy of commodity DRAMs is organized into channels, ranks, chips, banks, rows, and columns, respectively~\citep{Ref_Ghose_DRAMworkload_POMACS19, Ref_Liu_RAIDR_SIGARCH12, Ref_Putra_DRMap_DAC20, Ref_Putra_ROMANet_TVLSI21}; see Fig.~\ref{Fig_NCL_DRAM}(b).
DRAM banks are the lowest-level units that can be accessed in parallel, known as bank-level parallelism~\cite{kim2010thread}.
In this paper, the number of DRAM channels-per-system, ranks-per-channel, chips-per-rank, banks-per-chip, rows-per-bank, and columns-per-bank are represented as $D_{chn}$, $D_{rnk}$, $D_{chp}$, $D_{bnk}$, $D_{row}$, and $D_{col}$, respectively.  
When a DRAM request is issued from the host CPU, a specific rank will respond, enabling parallel access to multiple chips in that rank whose data form a DRAM word. 
In each chip, the request is routed to a specific bank with a specific pair of row and column addresses. 
Then, the requested read or write can be performed.

\section{The RobustNCL Framework}
\label{Sec_RobustNCL}

RobustNCL framework (Fig.~\ref{Fig_RobustNCL}) addresses the targeted problem through NCL transient-fault modeling, fault-tolerance analysis, detect-and-cure (D\&C) strategy, and curing weight selection, as discussed in the following subsections.

\subsection{NCL Transient-Fault Modeling}
\label{Sec_RobustNCL_Modeling}

\begin{figure}[t]
    \centering
    \includegraphics[width=\linewidth]{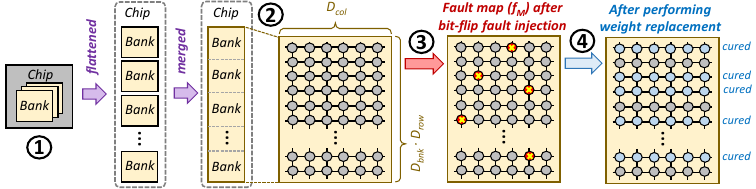}
    \caption{Transient-fault modeling in the DRAM with several key steps: flattening multiple banks into a continuous memory space, injecting transient faults, and curing anomalous weights. 
    Transient fault injection can happen at any time, as illustrated in Fig.~\ref{Fig_RobustNCL}.}
    \label{Fig_FaultModel}
\end{figure}

To properly investigate the impact of transient faults in the DRAM on NCL accuracy, \textit{we develop a novel NCL transient-fault modeling}, as shown in Fig.~\ref{Fig_FaultModel}.
Its key idea is to consider fault generation and distribution for each DRAM chip, as each chip can be deployed independently on the system platform; see~\circled{1}.
Multiple banks in the chip are considered as a continuous memory space with the dimension of $D_{mem} = (D_{bnk} \cdot D_{row}) \times D_{col}$ memory cells, where each cell stores a single bit of data; see~\circled{2}.
Afterward, transient faults are randomly injected or distributed into the DRAM memory cells in the form of bit-flips based on the given fault rate $f_R$, as transient faults typically occur in random locations on a chip~\citep{baumann2005radiation, putra2022softsnn}; see~\circled{3}. 
The fault rate $f_R$ is defined as the ratio between the faulty bits and the total number of memory cells $D_{mem}$.
Each faulty bit stays in the DRAM until it is overwritten by a new data bit; see~\circled{4}. 
Furthermore, since transient fault injection can happen at any time, this fault model considers injecting transient faults into the DRAM in different possible states of trained models. 
Consequently, the faults can affect the initial, offline-trained model (stage \circledB{1} in Fig.~\ref{Fig_RobustNCL}) or the latest updated NCL-trained model (stage \circledB{6} in Fig.~\ref{Fig_RobustNCL}).

\subsection{NCL Fault-Tolerance Analysis}
\label{Sec_RobustNCL_Analysis}

\begin{figure}[t]
    \centering
    \includegraphics[width=\linewidth]{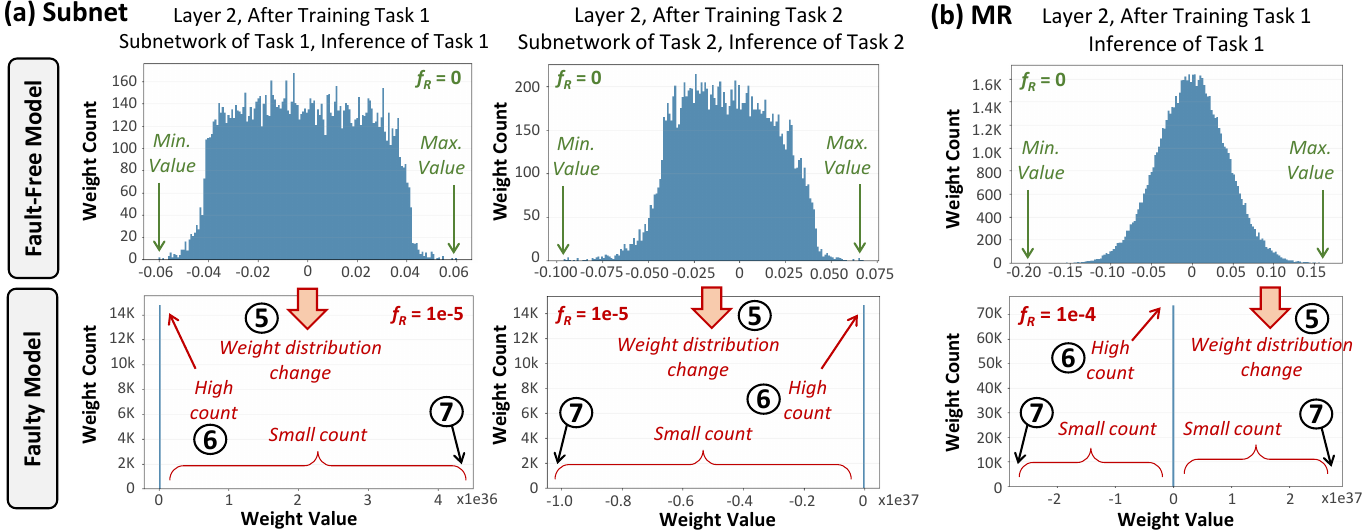}
    \caption{Experimental results from different NCL methods: (a) Subnet and (b) MR. 
    These results show weight distributions of both the fault-free and faulty models from their layer 2 at certain model states (i.e., after training task 1 and task 2), inference tasks (i.e., task 1 and task 2), and fault rates ($f_R = 1{\times}10^{-5}$ for Subnet and $f_R = 1{\times}10^{-4}$ for MR).}
    \label{Fig_WeightDist}
\end{figure}

This step aims to understand NCL fault tolerance and its characteristics through experimental case studies that investigate NCL accuracy under different fault rates $f_R$ of transient faults in the DRAM, while considering the following aspects.
\begin{itemize}
    \item State-of-the-art NCL models often use floating-point 32-bit (FP32) precision for their weights; thus, we also use the same FP32 precision format and set the flattened $D_{col} = 32$.      
    \item We consider a single DDR3-1600 2Gb DRAM chip to store the entire weights of the investigated NCL model, as this chip is widely used in many AI compute systems~\citep{Ref_Putra_ROMANet_TVLSI21, Ref_Putra_SparkXD_DAC21}.
    Since we set the flattened $D_{col} = 32$, then the flattened $D_{bnk} \cdot D_{row} = 2\text{Gb}/D_{col} = 2\text{Gb}/32$.
    \item Transient fault injection is performed after initial model deployment (stage \circledB{1} in Fig.~\ref{Fig_RobustNCL}) and after every NCL training phase for each new task (stage \circledB{6} in Fig.~\ref{Fig_RobustNCL}), while considering the fault modeling in Sec.~\ref{Sec_RobustNCL_Modeling}.
    \item We apply the Subnet and MR methods separately to the spiking-based VGG-9 architecture~\citep{apolinario2025tess} to provide comprehensive observations and insights across different NCL methods, while considering the Split MNIST benchmarking dataset. 
\end{itemize}
Details of the experimental setup are further discussed in Sec.~\ref{Sec_ExpSetup}.  
Some of the results from the experimental case studies are presented in Fig.~\ref{Fig_WeightDist}, from which we make the following key observations.
\begin{itemize}
    \item In general, bit-flips may change the weight distribution of each layer (or even each subnetwork) since they can notably change the weight values, depending on the faulty bit position within the selected precision format; see~\circled{5}.
    \item A bit-flip at the sign bit can switch the polarity of the weight value, hence affecting neural dynamics and output behavior.
    Specifically, a higher absolute value will lead to a more significant impact on neural dynamics and outputs.
    \item Some bit-flips may shift weights toward values that are very close to each other, so that they can be clustered in the same bin and notably increase the weight count; see~\circled{6}. 
    \item Bit-flips at significant bits of exponents may explode the weight values (i.e., either toward extreme positive or negative values), hence significantly deteriorate the neural dynamics; see~\circled{7}. 
    \item In the Subnet method, we observe a unique behavior.
    Transient faults in the DRAM may affect both the trained weights (from trained subnetworks) and the untrained weights (from untrained subnetworks). 
    Here, we observe that bit errors in untrained weights may lead to bad weight initialization, which in turn makes the corresponding subnetworks difficult to train.
    Therefore, we should mitigate this issue as well.
    \item In this step, we also record several important profiles of the weight distribution from each layer of the clean, fault-free model, including the minimum value ($w_{min}$), the maximum value ($w_{max}$), and the most-frequent/most-probable value ($w_{pro}$). 
\end{itemize}
All these key observations are then leveraged to develop fault mitigation techniques in Sec.~\ref{Sec_RobustNCL_DnC}. 

\subsection{Lightweight Detect-and-Cure (D\&C) Strategy}
\label{Sec_RobustNCL_DnC}

To detect and mitigate transient faults at runtime, our RobustNCL framework employs the Detect-and-Cure (D\&C) technique, which actively monitors whether anomalous weight values ($W^*$) are accessed from the DRAM, and proactively replaces (cures) them with safe values, thereby bounding the cured weight values ($W^C$) within a safe range that does not cause erroneous neural behavior. 

For a clear explanation, we discuss the proposed D\&C technique while describing the full stages of our RobustNCL framework below, based on Fig.~\ref{Fig_RobustNCL}. 

\textbf{Stage-\circledB{1} Initial Deployed Model:}
This stage represents the initial deployment of the offline-trained model ($M_1$).
Its clean, trained weights ($W$) are fully stored in the DRAM. 
If transient faults appear, then the clean weights $W$ will be corrupted into faulty weights ($W^*$). 
Here, the transient-fault modeling follows the discussion in Sec.~\ref{Sec_RobustNCL_Modeling}. 
To address these faulty weights $W^*$, our D\&C strategy in stage-\circledB{2} is employed.

\textbf{Stage-\circledB{2} D\&C Processing Strategy:}
This stage aims to cure faulty weights $W^*$ accessed from the DRAM.
Specifically, it performs our D\&C technique on each weight accessed from the DRAM ($\bar{w}$).
It first detects if $\bar{w}$ is outside the clean weight distribution (i.e., $\bar{w} > w_{max}$ or $\bar{w} < w_{min}$). 
If so, then it is considered a faulty weight ($\bar{w} \rightarrow w^*$) and will be replaced with the selected curing weight ($w^c$). 
As a result, all weights of the model are cured ($W^C$) and safe to use for both the inference and NCL training phases. 

\textbf{Stage-\circledB{3} Curing Weight Library:}
This stage provides weight candidates for each network layer to replace faulty weights, based on statistics ($w_{min}$, $w_{max}$, and $w_{pro}$) observed during the training phase. 
Here, there are two conditions.
\begin{itemize}
    \item In the initial deployment, this library only records statistics from the offline-trained model.
    Consequently, the selected curing value $w^c$ follows the initial statistics that are recorded in the offline fault-tolerance analysis in Sec.~\ref{Sec_RobustNCL_Analysis}. 
    \item As the model gets updated through NCL training at runtime, this library also updates its records based on updated statistics. 
    Consequently, the selected curing value $w^c$ follows the updated statistics in the NCL training phase in stage-\circledB{5}. 
\end{itemize}    

\textbf{\textit{Weight Replacement Variants}:}
To provide appropriate curing weight values, we propose different solution variants based on the recorded statistics of the per-layer weight distribution, including the minimum value $w_{min}$, the minimum value $w_{max}$, and the most-probable value $w_{pro}$. 
For the Subnet method in particular, we record statistics for each subnetwork per layer, as different subnetworks may have different weight distributions.
The proposed weight replacement variants include the following.
\begin{itemize}
    \item \textbf{RobustNCLv1 (Zero):} 
    If the DRAM-accessed weight $\bar{w}$ is anomalous, then it is considered faulty ($\bar{w} \rightarrow w^*$). 
    Then, we replace this faulty weight with zero, leading to $w^* \leftarrow w^c \; \text{with} \; w^c = 0$.
    \item \textbf{RobustNCLv2 (MinMax):} 
    If the DRAM-accessed weight $\bar{w}$ is anomalous, then it is considered faulty ($\bar{w} \rightarrow w^*$). 
    Then, we clip this faulty weight based on its value relative to the corresponding weight distribution. 
    If $w^* > w_{max}$, then we replace $w^*$ with the maximum value from the distribution, leading to $w^* \leftarrow w^c \; \text{with} \; w^c = w_{max}$.
    Meanwhile, if $w^* < w_{min}$, then we replace $w^*$ with the minimum value from the distribution, leading to $w^* \leftarrow w^c \; \text{with} \; w^c = w_{min}$.
    \item \textbf{RobustNCLv3 (Pro):} 
    If the DRAM-accessed weight $\bar{w}$ is anomalous, then it is considered faulty ($\bar{w} \rightarrow w^*$). 
    Then, we replace this faulty weight with the most-probable value, leading to $w^* \leftarrow w^c \; \text{with} \; w^c = w_{pro}$.
\end{itemize}

Since these curing weight values can also be affected by transient faults, we consider storing them in a radiation-hardened memory to ensure resistance against high-energy particle strikes. 
To achieve this, the existing hardening techniques that enhance the fabrication process~\citep{garg2009circuit, huang2019radhard} can be employed.

\textbf{Stage-\circledB{4} Inference Phase:}
It performs inference using the cured weights $W^C$ from the D\&C stage.

\textbf{Stage-\circledB{5} NCL Training Phase:}
This stage performs the NCL training to update the model with a new task ($T{+}1$).
Specifically, the previously trained weights ($W^C_{1-T}$) are updated by training on a new task ($T{+}1$), becoming the latest updated weights ($W^C_{1-(T{+}1)}$).
The recorded statistics from this phase are stored in the curing weight library, so they can be used to guide the D\&C processing stage. 

\textbf{Stage-\circledB{6} NCL-Trained Model:}
This stage stores the updated weights ($W_{1-(T{+}1)}$) in the DRAM, called as healthy weights ($\tilde{W}$).  
If transient faults appear, then these healthy weights $\tilde{W}$ will be corrupted into faulty weights ($W^*$). 
Again, the transient-fault modeling follows the discussion in Sec.~\ref{Sec_RobustNCL_Modeling}, and their faulty weights will be addressed by our D\&C strategy in stage-\circledB{2}.

\section{Experimental Setup}
\label{Sec_ExpSetup}

To evaluate our RobustNCL framework, we implement it in Python and run it on a single Nvidia RTX 4090 GPU device. 
It considers the TIL scenario, each benchmarking dataset is divided into a sequence of tasks: task 1, task 2, ..., task $K$; where $K$ denotes the maximum number of accommodated tasks. 
Task 1 is learned in the offline training phase to obtain the initial model to deploy for the first time; while the remaining tasks are learned sequentially in order during the NCL training phase.

\textbf{NCL Methods and Benchmarking Datasets:}
To show the generality of our RobustNCL framework, we consider: (1) the task-specific subnetwork (Subnet) method to represent the architecture-based approach, and (2) the memory replay (MR) method~\citep{minhas2025continual}.
The evaluation considers diverse benchmarking datasets, encompassing Split MNIST, Split Fashion MNIST (FMNIST), Split CIFAR-10, and Split CIFAR-100.
Each 5-task dataset (i.e., Split MNIST, Split FMNIST, and Split CIFAR-10) has 5 tasks, each having 2 classes-per-task. 
Meanwhile, the 10-task dataset (i.e., Split CIFAR-100) has 10 tasks with 10 classes-per-task. 
The evaluation outputs include accuracy, as well as processing time, memory, and power/energy consumption obtained from Nvidia NVML utilities.

\textbf{Network, Hardware, and Fault Rate Settings:}
We use the same network architecture for all experiments: the VGG9 network with 8 convolutional and 1 fully-connected layers~\citep{apolinario2025tess}. 
To support the TIL scenario, the last layer is structured so that each task is associated with a specific FC head. 
For the 5-task datasets (Split MNIST, Split Fashion MNIST, and Split CIFAR-10), each task training considers 5 timesteps, 8 epochs-per-task, and a learning rate of $\eta = 0.005$. 
For the 10-task dataset (Split CIFAR-100), each task training considers 9 timesteps, 100 epochs-per-task, and a learning rate of $\eta = 0.005$. 
Both Subnet and MR use a batch size of 32. 
Additionally, MR uses a buffer of 2,000 samples and a replay batch of 48. 
Furthermore, we consider a single DDR3-1600 2Gb DRAM chip~\citep{Ref_Micron} and different fault rates, as mentioned below.
 \begin{itemize}
     \item For Subnet on 5-task datasets: 
     $f_R \in \{5{\times}10^{-9}, 1{\times}10^{-8}, 5{\times}10^{-8}, 1{\times}10^{-7}, 5{\times}10^{-7}, 1{\times}10^{-6}, \\ 5{\times}10^{-6}, 1{\times}10^{-5} \}$.
     \item For MR on 5-task datasets: 
     $f_R \in \{1{\times}10^{-7}, 5{\times}10^{-7}, 1{\times}10^{-6}, 5{\times}10^{-6}, 1{\times}10^{-5}, 5{\times}10^{-5}, \\ 1{\times}10^{-4}, 5{\times}10^{-4} \}$.
     \item For both Subnet and MR on Split CIFAR-100: 
     $f_R \in \{1{\times}10^{-8}, 1{\times}10^{-7}, 1{\times}10^{-6}, 1{\times}10^{-5} \}$.
 \end{itemize}

\section{Results and Discussion}
\label{Sec_Results}

\subsection{Maintaining High Accuracy of NCL under Transient Faults}
Table~\ref{Tab_Subnet_5Tasks}, Table~\ref{Tab_MR_5Tasks}, and Table~\ref{Tab_10Tasks} present the experimental results on the final average accuracy (\%) evaluated across all trained tasks and across different fault-mitigation techniques for: Subnet with 5-task datasets, MR with 5-task datasets, and both Subnet and MR with 10-task datasets, respectively.

In Table~\ref{Tab_Subnet_5Tasks} (Subnet \& 5-task datasets), the baseline suffers from more than -10\% accuracy drop when encountering fault rates of $f_R \geq $ \rate{1}{7} for Split MNIST and Split CIFAR-10, and $f_R \geq $ \rate{5}{8} for Split FMNIST. 
In Table~\ref{Tab_MR_5Tasks} (MR \& 5-task datasets), the baseline again suffers from more than -10\% accuracy drop when encountering fault rates of $f_R \geq $ \rate{5}{5} for Split MNIST and Split CIFAR-10, and generally $f_R \geq $ \rate{1}{5} for Split FMNIST.
Similarly, in Table~\ref{Tab_10Tasks} (10-task datasets), the baseline suffers from more than -10\% accuracy drop when encountering fault rates of $f_R \geq $ \rate{1}{7} for Subnet, and $f_R \geq $ \rate{1}{6} for MR.
The reason is that, the baseline weights are very sensitive to bit-flips, especially at their significant bit locations, such as the sign and exponent bits, thereby leading to significant accuracy degradation if these bits are flipped, even at a low $f_R$.

In contrast, all of our RobustNCL variants successfully maintain NCL high accuracy of across different NCL methods, datasets, and fault rates. 
In Table~\ref{Tab_Subnet_5Tasks} (Subnet \& 5-task datasets), the best RobustNCL variant maintains no accuracy loss from the fault-free baseline, as they achieve final average accuracy between 99.79\% -- 99.82\% (Split MNIST), 99.01\% -- 99.04\% (Split FMNIST), and 84.88\% -- 85.25\% (Split CIFAR-10) even at high fault rate $f_R=$ \rate{1}{5}, with RobustNCLv2 (MinMax) always achieves the highest accuracy. 
In Table~\ref{Tab_MR_5Tasks} (MR \& 5-task datasets), the best RobustNCL variant maintains accuracy within -0.5\% from the fault-free baseline, as they achieve final average accuracy between 99.34\% -- 99.51\% (Split MNIST), 98.31\% -- 98.92\% (Split FMNIST), and 86.31\% -- 88.31\% (Split CIFAR-10) even at high fault rate $f_R=$ \rate{5}{4}.
Similarly, in Table~\ref{Tab_10Tasks} (10-task datasets), the best RobustNCL variant maintains no loss in accuracy from the fault-free baseline, as they achieve final average accuracy between 70.59\% -- 71.02\% (Subnet), 65.90\% -- 66.82\% (MR) at high fault rate $f_R=$ \rate{1}{5}.
All these improvements obtained by our RobustNCL variants are due to the D\&C strategy, which always monitors each DRAM-accessed weight to detect if it has an anomalous signature (i.e., too-small or too-big value) relative to the known clean/cured weight distributions for the corresponding layer or subnetwork-per-layer, and if so, it proactively cures this faulty weight accordingly.
Moreover, our updated curing weight library ensures the replacement/curing process remains relevant and updated with the latest NCL-trained model, thereby adapting to the dynamic nature of NCL model.  
Furthermore, we also observe that RobustNCLv2 (MinMax) generally performs slightly better than the other variants for 5-task datasets, while RobustNCLv3 (Pro) generally performs slightly better than the other variants for the 10-task dataset, at high fault rate (e.g., $f_R=$ \rate{1}{5}) and across different NCL methods. 
These indicate that those variants provide a better representation of the weight distributions for the given workload/dataset.

\begin{table}[h]
\footnotesize
\caption{Final average accuracy (\%) obtained by the Subnet method on the 5-task datasets evaluated across all trained tasks (task 1 -- task 5) and across different fault-mitigation techniques. 
}
\label{Tab_Subnet_5Tasks}
\begin{center}
\setlength{\tabcolsep}{2.4pt}
\begin{tabular}{ll*{8}{c}@{\hspace{10pt}}c@{\hspace{7pt}}c}
\toprule
Dataset & Technique & \rate{5}{9} & \rate{1}{8} & \rate{5}{8} & \rate{1}{7} & \rate{5}{7} & \rate{1}{6} & \rate{5}{6} & \rate{1}{5} \\ 
\midrule
\midrule
\multirow{4}{*}{\begin{tabular}[c]{@{}c@{}}Split\\MNIST\end{tabular}} & Baseline & \textbf{99.84} & \textbf{99.83} & 90.05 & \addx{76.38} & \addx{51.28} & \addx{50.32} & \addx{51.28} & \addx{51.28} \\
 & RobustNCLv1 & 99.81 & 99.78 & 99.79 & \textbf{99.66} & 99.72 & 99.77 & 99.77 & 99.79 \\ 
 & RobustNCLv2 & 99.81 & 99.78 & \textbf{99.80} & \textbf{99.66} & 99.74 & \textbf{99.79} & \textbf{99.80} & \add{\textbf{99.82}} \\ 
 & RobustNCLv3 & 99.81 & 99.78 & 99.79 & \textbf{99.66} & \textbf{99.76} & 99.76 & 99.72 & 99.80 \\ 
\midrule
\multirow{4}{*}{\begin{tabular}[c]{@{}c@{}}Split\\FMNIST\end{tabular}} & Baseline & \textbf{99.01} & 89.36 & \addx{66.79} & \addx{57.54} & \addx{50.00} & \addx{50.00} & \addx{50.00} & \addx{50.00} \\ 
 & RobustNCLv1 & \textbf{99.01} & 99.05 & 99.03 & \textbf{99.00} & 98.98 & 99.06 & \textbf{99.05} & 99.01 \\ 
& RobustNCLv2 & \textbf{99.01} & \textbf{99.06} & 99.03 & 98.95 & 99.02 & \textbf{99.09} & 99.04 & \add{\textbf{99.04}} \\ 
 & RobustNCLv3 & \textbf{99.01} & 99.05 & \textbf{99.04} & 98.96 & \textbf{99.07} & \textbf{99.09} & 98.99 & 99.01 \\ 
\midrule
\multirow{4}{*}{\begin{tabular}[c]{@{}c@{}}Split\\CIFAR-10\end{tabular}} & Baseline & 84.35 & 84.74 & 76.79 & \addx{70.39} & \addx{50.00} & \addx{50.00} & \addx{50.00} & \addx{50.00} \\ 
 & RobustNCLv1 & \textbf{84.85} & \textbf{85.17} & 84.89 & 85.42 & 85.02 & \textbf{85.17} & 84.92 & 85.05 \\ 
 & RobustNCLv2 & \textbf{84.85} & \textbf{85.17} & \textbf{84.92} & 85.42 & 85.10 & 84.78 & \textbf{85.71} & \add{\textbf{85.25}} \\ 
 & RobustNCLv3 & \textbf{84.85} & \textbf{85.17} & 84.88 & \textbf{85.43} & \textbf{85.23} & 84.92 & 85.19 & 84.88 \\ 
\bottomrule
\end{tabular}
\end{center}
\end{table}

\begin{table}[t]
\footnotesize
\caption{Final average accuracy (\%) obtained by the MR method on the 5-task datasets evaluated across all trained tasks (task 1 -- task 5) and across different fault-mitigation techniques.
}
\label{Tab_MR_5Tasks}
\begin{center}
\setlength{\tabcolsep}{2.4pt}
\begin{tabular}{ll*{8}{c}@{\hspace{10pt}}c@{\hspace{7pt}}c}
\toprule
Dataset & Technique & \rate{1}{7} & \rate{5}{7} & \rate{1}{6} & \rate{5}{6} & \rate{1}{5} & \rate{5}{5} & \rate{1}{4} & \rate{5}{4} \\ 
\midrule
\midrule
\multirow{4}{*}{\begin{tabular}[c]{@{}c@{}}Split\\MNIST\end{tabular}} & Baseline & \textbf{99.40} & \textbf{99.37} & 99.38 & \textbf{99.47} & 98.61 & \addx{51.50} & \addx{51.50} & \addx{51.50} \\ 
 & RobustNCLv1 & 99.33 & 99.31 & 99.41 & 99.22 & 99.40 & \textbf{99.41} & 99.37 & 99.34 & \\ 
 & RobustNCLv2 & 99.39 & 99.35 & 99.19 & 99.44 & \textbf{99.49} & 99.30 & 99.28 & \add{\textbf{99.51}} \\ 
 & RobustNCLv3 & 99.29 & 99.32 & \textbf{99.49} & 99.38 & 99.40 & 99.23 & \textbf{99.39} & 99.37 \\ 
\midrule
\multirow{4}{*}{\begin{tabular}[c]{@{}c@{}}Split\\FMNIST\end{tabular}} & Baseline & 99.03 & 98.99 & \textbf{99.08} & 98.89 & \addx{87.37} & 94.14 & \addx{50.00} & \addx{50.00} \\ 
 & \curezero & 99.01 & 98.97 & 99.07 & 98.96 & 99.00 & 99.06 & \textbf{99.01} & 98.31 \\ 
 & \curemm & \textbf{99.07} & \textbf{99.06} & 99.04 & \textbf{99.06} & \textbf{99.01} & 99.07 & 98.96 & \add{\textbf{98.92}} \\
 & \curemode & 99.06 & 98.98 & 99.01 & 99.01 & 98.99 & \textbf{99.08} & \textbf{99.01} & 98.81 \\
\midrule
\multirow{4}{*}{\begin{tabular}[c]{@{}c@{}}Split\\CIFAR-10\end{tabular}} & Baseline & \textbf{88.80} & 88.39 & 88.49 & 86.53 & 87.85 & \addx{50.00} & \addx{50.00} & \addx{50.00} \\ 
 & \curezero & 88.31 & \textbf{88.76} & 88.10 & 88.60 & 87.31 & 88.37 & 87.64 & 86.31 \\ 
 & \curemm & 88.03 & 88.59 & \textbf{88.58} & 88.53 & 88.32 & \textbf{88.98} & \textbf{88.15} & 86.37 \\
 & \curemode & 87.61 & 88.12 & 88.47 & \textbf{88.67} & \textbf{88.56} & 87.85 & 87.07 & \add{\textbf{88.31}} \\
\bottomrule
\end{tabular}
\end{center}
\end{table}

\begin{table}[t]
\footnotesize 
\caption{Final average accuracy (\%) obtained by both the Subnet and MR methods on the 10-task dataset (Split CIFAR-100) evaluated across all trained tasks (task 1 -- task 10).
}
\label{Tab_10Tasks}
\begin{center}
\setlength{\tabcolsep}{2.4pt}
\begin{tabular}{ll*{8}{c}@{\hspace{10pt}}c@{\hspace{7pt}}c}
\toprule
\multirow{2}{*}{\begin{tabular}[c]{@{}c@{}}Technique\end{tabular}} & \multicolumn{4}{|c}{Subnet} & \multicolumn{4}{|c}{MR}\\
&  \multicolumn{1}{|c}{\rate{1}{8}} & \rate{1}{7} & \rate{1}{6} & \rate{1}{5} & \multicolumn{1}{|c}{\rate{1}{8}} & \rate{1}{7} & \rate{1}{6} & \rate{1}{5} \\  
\midrule
\midrule
Baseline & 60.00 & \addx{10.00} & \addx{10.71} & \addx{10.00} & 66.66 & 66.62 & \addx{10.00} & \addx{10.00} \\
RobustNCLv1 & \textbf{70.73} & 70.61 & 70.61 & 70.82 & \textbf{67.11} & \textbf{66.91} & \textbf{66.84} & 66.35 \\ 
RobustNCLv2 & \textbf{70.73} & \textbf{70.66} & 70.50 & 70.59 & 66.75 & 66.35 & 66.59 & 65.90 \\ 
RobustNCLv3 & \textbf{70.73} & 70.57 & \textbf{70.66} & \add{\textbf{71.02}} & 66.65 & 66.63 & 66.25 & \add{\textbf{66.82}} \\  
\bottomrule
\end{tabular}
\end{center}
\end{table}

\subsection{Overheads of RobustNCL}
RobustNCL incurs negligible processing time and memory overheads as compared to the baseline without any fault mitigation (i.e., up to 2.3\% latency overheads and less than 1\% memory overheads) across different NCL methods, datasets, and fault rates. 
This is because RobustNCL only requires small additional time and memory for performing the D\&C process for each weight accessed from the DRAM, and for updating the curing weight library after the NCL training phase.   
RobustNCL also incurs a small overhead of energy consumption (i.e., 6.35\% on average) across different NCL methods, datasets, and fault rates.   
The reason is that RobustNCL incurs moderate additional power to perform the D\&C process and to update the curing weight library after the NCL training phase.
In summary, these overhead results and analysis highlight that our RobustNCL framework successfully offers a low-cost fault-mitigation solution against hardware-level transient faults in the DRAM across different NCL methods, datasets, and fault rates, showcasing its good generality and scalability.

\section{Conclusion}
\label{Sec_Conclude}

In this paper, we propose a novel RobustNCL framework to improve NCL reliability in the presence of transient faults in the DRAM off-chip/main memory through a lightweight detect-and-cure strategy. 
It detects faults during weight data access from the DRAM, then cures them through three curing variants: replacement with zero, min/max value, and most-probable value from the clean/cured weight distribution.
RobustNCL successfully preserves high accuracy scores under high fault rates close to those of fault-free models  (i.e., within -0.5\% accuracy for the best RobustNCL variant) and significantly improves accuracy from the baseline without fault mitigation across different NCL methods and benchmarking datasets, while incurring negligible latency and energy overheads. 
These results showcase that our RobustNCL framework is a promising solution to realize reliable and energy-efficient NCL systems.

\bibliography{iclr2027_conference}

@inproceedings{kim2010thread,
  title={Thread cluster memory scheduling: Exploiting differences in memory access behavior},
  author={Kim, Yoongu and Papamichael, Michael and Mutlu, Onur and Harchol-Balter, Mor},
  booktitle={2010 43rd Annual IEEE/ACM International Symposium on Microarchitecture},
  pages={65--76},
  year={2010},
  organization={IEEE}
}

@article{garg2009circuit,
  title={Circuit-level design approaches for radiation-hard digital electronics},
  author={Garg, Rajesh and Jayakumar, Nikhil and Khatri, Sunil P and Choi, Gwan S},
  journal={IEEE transactions on very large scale integration (VLSI) systems},
  volume={17},
  number={6},
  pages={781--792},
  year={2009},
  publisher={IEEE}
}

@article{huang2019radhard,
  title={An overview of radiation effects on electronic devices under severe accident conditions in NPPs, rad-hardened design techniques and simulation tools},
  author={Huang, Qiang and Jiang, Jin},
  journal={Progress in Nuclear Energy},
  volume={114},
  pages={105--120},
  year={2019},
  publisher={Elsevier}
}

@article{delange2022continual,
  author  = {De Lange, Matthias and Aljundi, Rahaf and Masana, Marc and Parisot, Sarah and Jia, Xu and Leonardis, Ale{\v{s}} and Slabaugh, Gregory and Tuytelaars, Tinne},
  title   = {A Continual Learning Survey: Defying Forgetting in Classification Tasks},
  journal = {IEEE Transactions on Pattern Analysis and Machine Intelligence},
  volume  = {44},
  number  = {7},
  pages   = {3366--3385},
  year    = {2022},
  doi     = {10.1109/TPAMI.2021.3057446}
}

@article{wang2024comprehensive,
  author  = {Wang, Liyuan and Zhang, Xingxing and Su, Hang and Zhu, Jun},
  title   = {A Comprehensive Survey of Continual Learning: Theory, Method and Application},
  journal = {IEEE Transactions on Pattern Analysis and Machine Intelligence},
  volume  = {46},
  number  = {8},
  pages   = {5362--5383},
  year    = {2024},
  doi     = {10.1109/TPAMI.2024.3367329}
}

@inproceedings{lopezpaz2017gradient,
  author    = {Lopez-Paz, David and Ranzato, Marc'Aurelio},
  title     = {Gradient Episodic Memory for Continual Learning},
  booktitle = {Advances in Neural Information Processing Systems (NeurIPS)},
  volume    = {30},
  pages     = {6467--6476},
  year      = {2017}
}

@article{maass1997networks,
  author  = {Maass, Wolfgang},
  title   = {Networks of Spiking Neurons: The Third Generation of Neural Network Models},
  journal = {Neural Networks},
  volume  = {10},
  number  = {9},
  pages   = {1659--1671},
  year    = {1997},
  doi     = {10.1016/S0893-6080(97)00011-7}
}

@article{roy2019towards,
  author  = {Roy, Kaushik and Jaiswal, Akhilesh and Panda, Priyadarshini},
  title   = {Towards Spike-Based Machine Intelligence with Neuromorphic Computing},
  journal = {Nature},
  volume  = {575},
  number  = {7784},
  pages   = {607--617},
  year    = {2019},
  doi     = {10.1038/s41586-019-1677-2}
}

@article{davies2018loihi,
  author  = {Davies, Mike and Srinivasa, Narayan and Lin, Tsung-Han and Chinya, Gautham and Cao, Yongqiang and Choday, Sri Harsha and Dimou, Georgios and Joshi, Prasad and others},
  title   = {Loihi: A Neuromorphic Manycore Processor with On-Chip Learning},
  journal = {IEEE Micro},
  volume  = {38},
  number  = {1},
  pages   = {82--99},
  year    = {2018},
  doi     = {10.1109/MM.2018.112130359}
}

@inproceedings{apolinario2025tess,
  author    = {Apolinario, Marco P. E. and Roy, Kaushik and Frenkel, Charlotte},
  title     = {{TESS}: A Scalable Temporally and Spatially Local Learning Rule for Spiking Neural Networks},
  booktitle = {International Joint Conference on Neural Networks (IJCNN)},
  pages     = {1--9},
  year      = {2025},
  doi       = {10.1109/IJCNN64981.2025.11227652}
}

@article{minhas2025continual,
  author  = {Minhas, Mishal Fatima and Putra, Rachmad Vidya Wicaksana and Awwad, Falah and Hasan, Osman and Shafique, Muhammad},
  title   = {Continual Learning with Neuromorphic Computing: Foundations, Methods, and Emerging Applications},
  journal = {IEEE Access},
  volume  = {13},
  pages   = {124824--124873},
  year    = {2025},
  doi     = {10.1109/ACCESS.2025.3588665}
}

@article{antonov2022continuous,
  author  = {Antonov, Dmitry and Sviatov, Kirill and Sukhov, Sergey},
  title   = {Continuous Learning of Spiking Networks Trained with Local Rules},
  journal = {Neural Networks},
  volume  = {155},
  pages   = {512--522},
  year    = {2022},
  doi     = {10.1016/j.neunet.2022.09.003}
}

@article{skatchkovsky2022bayesian,
  author  = {Skatchkovsky, Nicolas and Jang, Hyeryung and Simeone, Osvaldo},
  title   = {Bayesian Continual Learning via Spiking Neural Networks},
  journal = {Frontiers in Computational Neuroscience},
  volume  = {16},
  pages   = {1037976},
  year    = {2022},
  doi     = {10.3389/fncom.2022.1037976}
}

@inproceedings{xiao2024hebbian,
  author    = {Xiao, Mingqing and Meng, Qingyan and Zhang, Zongpeng and He, Di and Lin, Zhouchen},
  title     = {Hebbian Learning Based Orthogonal Projection for Continual Learning of Spiking Neural Networks},
  booktitle = {International Conference on Learning Representations (ICLR)},
  year      = {2024},
  url       = {https://openreview.net/forum?id=MeB86edZ1P}
}

@inproceedings{han2023dynamic,
  author    = {Han, Bing and Zhao, Feifei and Zeng, Yi and Pan, Wenxuan and Shen, Guobin},
  title     = {Enhancing Efficient Continual Learning with Dynamic Structure Development of Spiking Neural Networks},
  booktitle = {International Joint Conference on Artificial Intelligence (IJCAI)},
  pages     = {2993--3001},
  year      = {2023},
  doi       = {10.24963/ijcai.2023/334}
}

@inproceedings{pes2024active,
  author    = {Pes, Lorenzo and Luiken, Rick and Corradi, Federico and Frenkel, Charlotte},
  title     = {Active Dendrites Enable Efficient Continual Learning in Time-to-First-Spike Neural Networks},
  booktitle = {IEEE 6th International Conference on AI Circuits and Systems (AICAS)},
  pages     = {41--45},
  year      = {2024},
  doi       = {10.1109/AICAS59952.2024.10595872}
}

@inproceedings{proietti2023memory,
  author    = {Proietti, Michela and Ragno, Alessio and Capobianco, Roberto},
  title     = {Memory Replay for Continual Learning with Spiking Neural Networks},
  booktitle = {IEEE 33rd International Workshop on Machine Learning for Signal Processing (MLSP)},
  pages     = {1--6},
  year      = {2023},
  doi       = {10.1109/MLSP55844.2023.10285911}
}

@inproceedings{dequino2024compressed,
  author    = {Dequino, Alberto and Carpegna, Alessio and Nadalini, Davide and Savino, Alessandro and Benini, Luca and Di Carlo, Stefano and Conti, Francesco},
  title     = {Compressed Latent Replays for Lightweight Continual Learning on Spiking Neural Networks},
  booktitle = {IEEE Computer Society Annual Symposium on VLSI (ISVLSI)},
  pages     = {240--245},
  year      = {2024},
  doi       = {10.1109/ISVLSI61997.2024.00052}
}

@inproceedings{minhas2025replay4ncl,
  author    = {Minhas, Mishal Fatima and Putra, Rachmad Vidya Wicaksana and Awwad, Falah and Hasan, Osman and Shafique, Muhammad},
  title     = {{Replay4NCL}: An Efficient Memory Replay-Based Methodology for Neuromorphic Continual Learning in Embedded {AI} Systems},
  booktitle = {62nd ACM/IEEE Design Automation Conference (DAC)},
  pages     = {1--7},
  year      = {2025},
  doi       = {10.1109/DAC63849.2025.11132839}
}

@article{baumann2005radiation,
  author  = {Baumann, Robert C.},
  title   = {Radiation-Induced Soft Errors in Advanced Semiconductor Technologies},
  journal = {IEEE Transactions on Device and Materials Reliability},
  volume  = {5},
  number  = {3},
  pages   = {305--316},
  year    = {2005},
  doi     = {10.1109/TDMR.2005.853449}
}

@inproceedings{hoang2020ftclipact,
  author    = {Hoang, Le-Ha and Hanif, Muhammad Abdullah and Shafique, Muhammad},
  title     = {{FT-ClipAct}: Resilience Analysis of Deep Neural Networks and Improving Their Fault Tolerance Using Clipped Activation},
  booktitle = {Design, Automation \& Test in Europe Conference \& Exhibition (DATE)},
  pages     = {1241--1246},
  year      = {2020},
  doi       = {10.23919/DATE48585.2020.9116571}
}

@inproceedings{putra2022softsnn,
  author    = {Putra, Rachmad Vidya Wicaksana and Hanif, Muhammad Abdullah and Shafique, Muhammad},
  title     = {{SoftSNN}: Low-Cost Fault Tolerance for Spiking Neural Network Accelerators under Soft Errors},
  booktitle = {59th ACM/IEEE Design Automation Conference (DAC)},
  pages     = {151--156},
  year      = {2022},
  doi       = {10.1145/3489517.3530657}
}

@article{Ref_Shaheen_CLsurvey_JIRS22,
  title={Continual learning for real-world autonomous systems: Algorithms, challenges and frameworks},
  author={Shaheen, Khadija and Hanif, Muhammad Abdullah and Hasan, Osman and Shafique, Muhammad},
  journal={Journal of Intelligent \& Robotic Systems},
  volume={105},
  number={1},
  pages={9},
  year={2022},
  publisher={Springer}
}

@INPROCEEDINGS{Ref_Putra_SpikeDyn_DAC21,
  author={Putra, Rachmad Vidya Wicaksana and Shafique, Muhammad},
  booktitle={58th ACM/IEEE Design Automation Conference (DAC)}, 
  title={SpikeDyn: A Framework for Energy-Efficient Spiking Neural Networks with Continual and Unsupervised Learning Capabilities in Dynamic Environments}, 
  year={2021},
  volume={},
  number={},
  pages={1057-1062},
  doi={10.1109/DAC18074.2021.9586281}}

@inproceedings{Ref_Putra_lpSpikeCon_IJCNN22,
  title={lpspikecon: Enabling low-precision spiking neural network processing for efficient unsupervised continual learning on autonomous agents},
  author={Putra, Rachmad Vidya Wicaksana and Shafique, Muhammad},
  booktitle={2022 International Joint Conference on Neural Networks (IJCNN)},
  pages={1--8},
  year={2022},
  organization={IEEE}
}

@ARTICLE{Ref_Putra_ROMANet_TVLSI21,
  author={Putra, Rachmad Vidya Wicaksana and Hanif, Muhammad Abdullah and Shafique, Muhammad},
  journal={IEEE Transactions on Very Large Scale Integration (VLSI) Systems},
  title={ROMANet: Fine-Grained Reuse-Driven Off-Chip Memory Access Management and Data Organization for Deep Neural Network Accelerators}, 
  year={2021},
  volume={29},
  number={4},
  pages={702-715},
  doi={10.1109/TVLSI.2021.3060509}}

@INPROCEEDINGS{Ref_Vadlamani_DMR_DATE10,
  title={Multicore soft error rate stabilization using adaptive dual modular redundancy},
  author={Vadlamani, Ramakrishna and Zhao, Jia and Burleson, Wayne and Tessier, Russell},
  booktitle={2010 Design, Automation \& Test in Europe Conference \& Exhibition (DATE 2010)},
  pages={27--32},
  year={2010},
  organization={IEEE}
}

@article{Ref_Lyons_TMR_IBM62,
author={Lyons, Robert E and Vanderkulk, Wouter},
title = {The Use of Triple-Modular Redundancy to Improve Computer Reliability},
year = {1962},
issue_date = {April 1962},
publisher = {IBM Corp.},
address = {USA},
volume = {6},
number = {2},
pages = {200–209},
issn = {0018-8646},
journal={IBM Journal of Research and Development},
numpages = {10}
}

@misc{Ref_Sze_ECCs_USPatent00,
  title={Circuit and method for rapid checking of error correction codes using cyclic redundancy check},
  author={Sze, Hung Y.},
  year={2000},
  month=jul # "~18",
  publisher={Google Patents},
  note={US Patent 6,092,231}
}

@article{Ref_Ghose_DRAMworkload_POMACS19,
  title={Demystifying complex workload-dram interactions: An experimental study},
  author={Ghose, Saugata and Li, Tianshi and Hajinazar, Nastaran and Cali, Damla Senol and Mutlu, Onur},
  journal={Proceedings of the ACM on Measurement and Analysis of Computing Systems},
  volume={3},
  number={3},
  pages={1--50},
  year={2019},
  publisher={ACM New York, NY, USA}
}

@article{Ref_Liu_RAIDR_SIGARCH12,
  title={RAIDR: Retention-aware intelligent DRAM refresh},
  author={Liu, Jamie and Jaiyen, Ben and Veras, Richard and Mutlu, Onur},
  journal={ACM SIGARCH Computer Architecture News},
  volume={40},
  number={3},
  pages={1--12},
  year={2012},
  publisher={ACM New York, NY, USA}
}

@article{Ref_Micron,
  author    = {Micron},
  title     = {Micron 2Gb: x4, x8, x16 DDR3 SDRAM. Datasheet MT41J128M16HA-12},
  year      = {2010},
}

@INPROCEEDINGS{Ref_Putra_DRMap_DAC20,
  title={DRMap: A generic DRAM data mapping policy for energy-efficient processing of convolutional neural networks},
  author={Putra, Rachmad Vidya Wicaksana and Hanif, Muhammad Abdullah and Shafique, Muhammad},
  booktitle={2020 57th ACM/IEEE Design Automation Conference (DAC)},
  pages={1--6},
  year={2020},
  organization={IEEE}
}

@INPROCEEDINGS{Ref_Putra_SparkXD_DAC21,
  author={Putra, Rachmad Vidya Wicaksana and Hanif, Muhammad Abdullah and Shafique, Muhammad},
  booktitle={58th ACM/IEEE Design Automation Conference (DAC)},
  title={SparkXD: A Framework for Resilient and Energy-Efficient Spiking Neural Network Inference using Approximate DRAM}, 
  year={2021},
  volume={},
  number={},
  pages={379-384},
  }

@article{Ref_LeCun_DeepLearning_Nature15,
  title={Deep learning},
  author={LeCun, Yann and Bengio, Yoshua and Hinton, Geoffrey},
  journal={Nature},
  volume={521},
  number={7553},
  pages={436--444},
  year={2015},
  publisher={Nature Publishing Group UK London}
}
\bibliographystyle{iclr2027_conference}

\end{document}